%% file: main.tex
\documentclass[runningheads]{llncs}

\usepackage{eccv}

\usepackage{eccvabbrv}
\usepackage[table]{xcolor}
\usepackage{xcolor}   
\usepackage{graphicx}
\usepackage{booktabs}
\usepackage{algorithm}
\usepackage{algorithmic}
\usepackage{amsmath}
\usepackage{amssymb}
\usepackage[accsupp]{axessibility}  % Improves PDF readability for those with disabilities.
\usepackage{marvosym}

\usepackage{hyperref}
\usepackage{pifont}

\usepackage{amsthm} 

\usepackage{orcidlink}
\usepackage{adjustbox}

\begin{document}

\title{Domain Adaptive Object Detection via Dual-Stream Bilevel-Cycle Optimization} 

\titlerunning{DAOD via Dual-Stream Bilevel-Cycle Optimization}

\author{%
{\fontsize{10.2}{12}\selectfont
Yannan~Chen\inst{1,4}\orcidlink{0009-0007-1896-3568} \and
Wei~Wang\inst{1}\textsuperscript{\Letter}\orcidlink{0000-0001-8676-1190} \and
Wenqiang~Wang\inst{1}\orcidlink{0009-0008-5034-1379} \and
Ruoyu~Chen\inst{2}\orcidlink{0000-0001-7630-7154}\\[-0.2em]
\makebox[\textwidth][c]{%
Jiancheng~Wang\inst{3}\orcidlink{0009-0005-8478-8177} \and
Mingbo~Yang\inst{1}\orcidlink{0009-0000-2952-5044} \and
Yaowei~Wang\inst{4}\orcidlink{0000-0003-2197-9038} \and
Xiaochun~Cao\inst{1}\textsuperscript{\Letter}\orcidlink{0000-0001-7141-708X}
}%
}%
}

\authorrunning{Y.~Chen et al.}

\institute{
School of Cyber Science and Technology, Sun Yat-sen University, \\Shenzhen Campus, Guangdong 518107, China
\and
University of Chinese Academy of Sciences, Beijing 100049, China
\and
School of Computer Science and Technology, Anhui University,\\ Hefei, Anhui 230601, China
\and
PC Laboratory, Shenzhen, Guangdong 518055, China
}
\maketitle

\begingroup
\renewcommand{\thefootnote}{\Letter}
\renewcommand{\theHfootnote}{corresponding-author-note}
\footnotetext[1]{Corresponding authors: Wei Wang and Xiaochun Cao. \\
Emails: \href{mailto:wangwei29@mail.sysu.edu.cn}{wangwei29@mail.sysu.edu.cn} and
\href{mailto:caoxiaochun@mail.sysu.edu.cn}{caoxiaochun@mail.sysu.edu.cn}.}
\endgroup

\input{sec/0_abstract}

\input{sec/1_introduction}
\input{sec/2_related_work}

\input{sec/3_methodology}

\input{sec/4_experiments}
\input{sec/5_conclusion}

\section*{Acknowledgements}
This work was supported in part by the National Natural Science Foundation of China
(Grant Nos. 62306343, U2541229, and 62441619), the Guangdong Basic and Applied Basic 
Research Foundation (Grant No. 2025A1515011322), and the Shenzhen Science and 
Technology Program (Grant Nos. KQTD20221101093559018 and SYSRD20250529113401002).
% Please insert your acknowledgments here.

% \bibliography{main}
% \newpage
% \input{sec/X_supplement}
% \newpage
\bibliographystyle{splncs04}
\bibliography{main}
\end{document}

%% file: sec/0_abstract.tex
\begin{abstract}
Cycle self-training (CST) breaks the shared classifier assumption of the standard self-training framework, which is effective for unsupervised domain adaptation and exploits unlabeled target data by training with target pseudo-labels. CST introduces a target classifier and employs an inner-outer loop updating strategy, addressing the issue of unreliable pseudo-labels and enabling pseudo-labels to generalize across domains. Despite its success in image classification, extending CST to object detection faces three main challenges. First, the upper bound of CST in object detection is constrained by three types of unreliable pseudo-labels, such as classification error alone, localization error alone, and their combination. Second, since object detection involves detecting multiple target objects, directly applying CST leads to training instability. Third, a wider numerical range of regression coordinates leads to exploding losses. To this end, we apply CST to both classification and regression and propose the Dual-Stream Bilevel-Cycle Optimization framework. Specifically, we construct CST upon Mean Teacher to prevent training instability and use extra normalization to map the regression bounding box into a standardized space, effectively addressing exploding losses. Also, we provide a theoretical derivation of the regression bound. Extensive experiments across four cross domain standard scenarios demonstrate that our framework achieves considerable results.
\keywords{Domain Adaptive Object Detection \and Self-training \and Mean Teacher}
\end{abstract}

%% file: sec/1_introduction.tex
\section{Introduction}
\label{sec:intro}
Self-Training (ST) serves as an effective paradigm in unsupervised domain adaptation \cite{liu21cycle}, which leverages a labeled source domain to annotate an unlabeled target domain, thereby reducing the high cost of manual annotation \cite{Amini2025Self}. Specifically, ST first utilizes a model trained on the source domain to generate pseudo-labels for the target domain \cite{Zhou26}. Subsequently, it iteratively retrains the model using both the ground-truth labels from the source domain and the generated pseudo-labels from the target domain \cite{yoon24enhancing,chen2025mgcamt,Wang2025Optimal}. However, due to the distribution shift between the source and target domains, the pseudo-labels for the target domain are often unreliable, \textit{i.e.}, they deviate from the ground-truth labels. To address this issue, Cycle Self-Training (CST) relaxes the shared classifier assumption of ST by introducing a specific target domain classifier and employing an inner-outer loop update strategy to mitigate the unreliability of pseudo-labels \cite{Li26,qu25cfnet}.

\begin{figure*}[t] 
\centering
\includegraphics[width=1.0\linewidth, height=0.38\textheight]{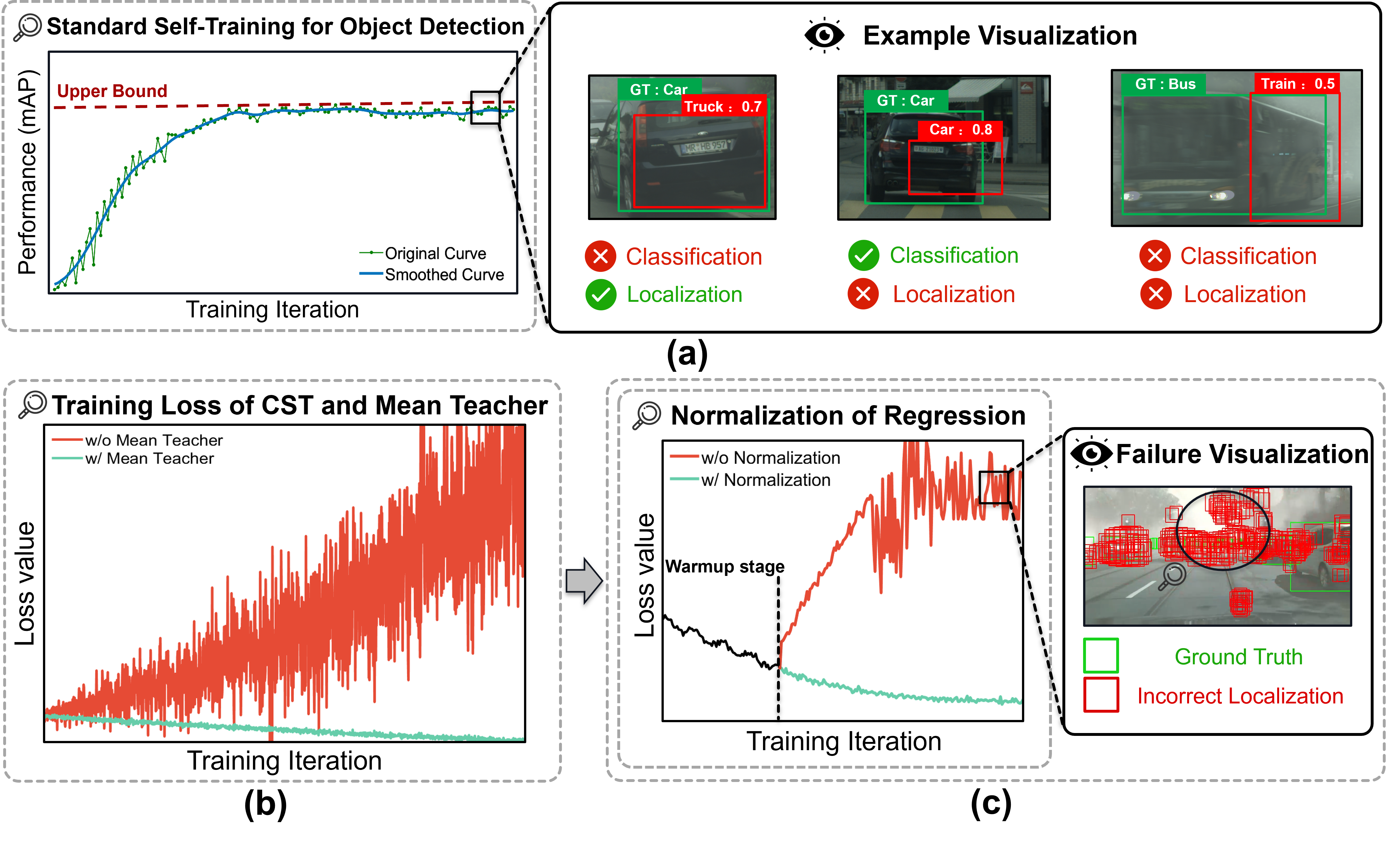} 
\caption{Motivation of this paper illustrating the challenges of extending CST to object detection. (a) Through empirical analysis, we observe that standard ST exhibits a performance upper bound in object detection. The visualization reveals that this limitation stems from incorrect classification, inaccurate localization, and combined failure. (b) Directly applying CST leads to training instability. (c) Implementing MT without regression normalization results in severe loss and model collapse where the visualization displays large bounding box shifts.}
\label{fig:motivation}
\end{figure*}

Although CST has achieved superior performance in classification tasks \cite{liu21cycle,yao25sfod}, its application in object detection remains underexplored. Through extensive experiments and theoretical analysis, we identify three primary challenges: \ding{182} As shown in Fig. ~\ref{fig:motivation}~(a), similar to classification tasks
\cite{li24csja, Wang2025Optimal}, ST in object detection also suffers from a performance upper bound. However, the unreliability of pseudo-labels in detection manifests in three forms: inaccurate classification, inaccurate localization, and inaccuracy in both classification and localization. \ding{183} As illustrated in Fig.~\ref{fig:motivation}~(b), directly applying CST to the Fully Convolutional One-Stage (FCOS) detector leads to training instability \cite{tian19fcos}. \ding{184} As depicted in Fig.~\ref{fig:motivation}~(c), Mean Teacher (MT) without normalization leads to severe bounding box shifts and model collapse \cite{Tarvainen18, Cao2023CMT, Weng2024MTM}. Therefore, despite its success in image classification, extending CST to object detection is still challenging \cite{krishna23mila,wang24adt, Chen2025Generalized}.

\begin{figure*}[t]
    \centering
    \includegraphics[width=1.0\linewidth]{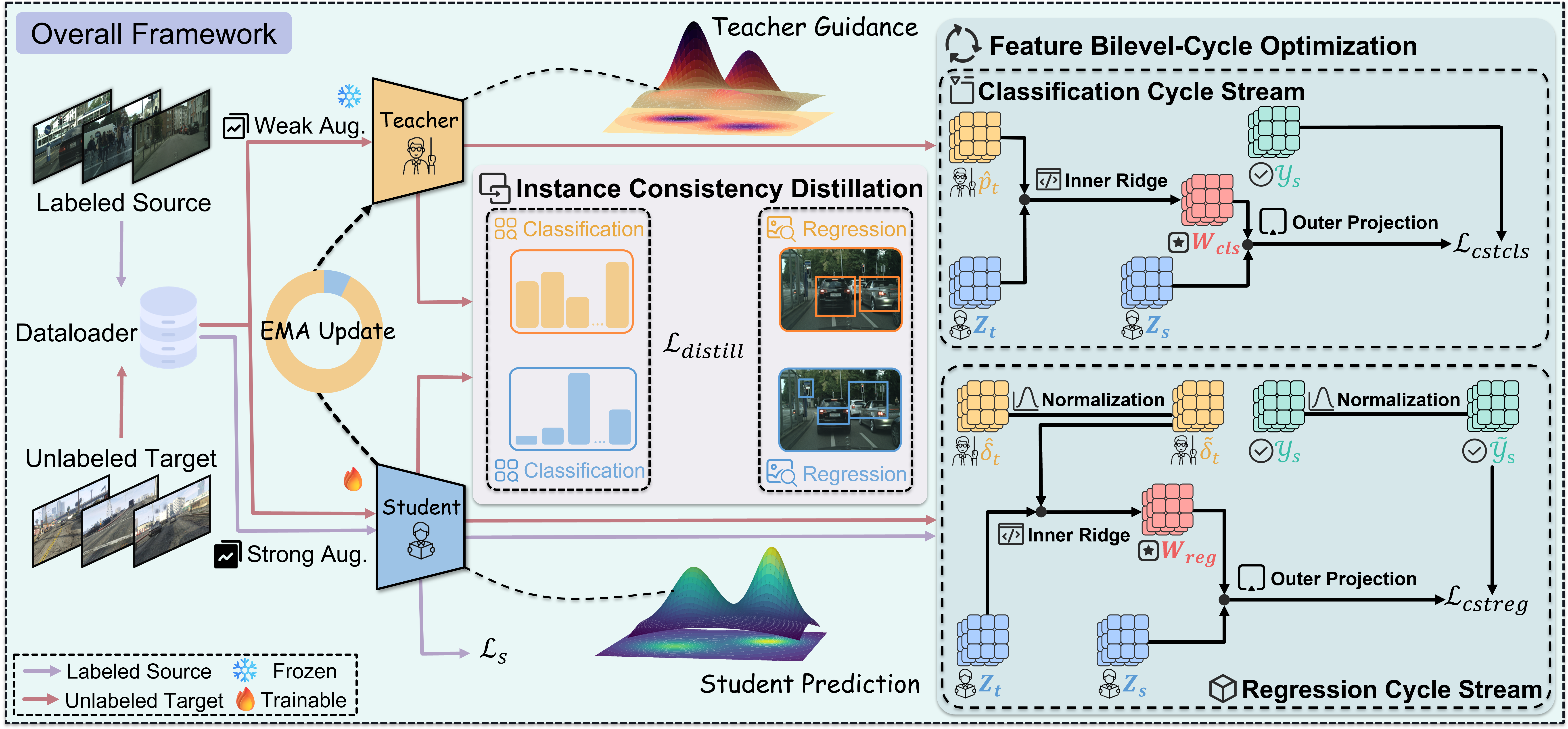}
   \caption{Overview of the proposed DSBCO framework constructed upon the stable MT paradigm. On the left, the trainable Student network processes strongly augmented images from both domains, while the frozen Teacher, updated via EMA, uses only weakly augmented target images to generate pseudo-labels. In the middle, Instance Consistency Distillation applies self-training as a baseline. On the right, Feature Bilevel-Cycle Optimization employs Classification and Regression Cycle Streams, where the latter introduces normalization to prevent exploding losses. In both streams, the Inner Ridge and Outer Projection work together to align the domains by mapping target features back to the source, effectively extending CST to the object detection task.} \label{fig:framework}
     % \vspace{-3.5em}
\end{figure*}

To break the performance upper bound caused by unreliable pseudo-labels (\textbf{Challenge \ding{182}}), we propose a \textbf{D}ual-\textbf{S}tream \textbf{B}ilevel-\textbf{C}ycle \textbf{O}p\-ti\-mi\-za\-tion (\textbf{DSB- CO}) framework, as shown in Fig.~\ref{fig:framework}. In DSBCO, we extend the CST mechanism to both classification and regression tasks. By enforcing consistency across domains, our DSBCO effectively filters out unreliable pseudo-labels. We address the training instability (\textbf{Challenge \ding{183}}) by constructing our framework upon the stable MT paradigm. Specifically, we leverage the exponential moving average (EMA) \cite{cai21eman} to smooth the parameter updates and stabilize the learning process in dense detection tasks.
Finally, to prevent model collapse caused by the wide numerical range of regression coordinates (\textbf{Challenge \ding{184}}), we introduce the regression normalization mechanism. This strategy projects the large regression targets into a standard distribution, which ensures stable gradient descent and effectively mitigates large-scale bounding box shifts.
 % results and findings
 
We validate the effectiveness of the proposed DSBCO framework through extensive experiments across four domain adaptation scenarios. Our method exceeds all existing FCOS detector methods. Specifically, compared to the second-best Harmonious Teacher (HT) \cite{deng2023harmonious}, DSBCO achieves effective improvements, increasing performance by 14.3\% in Adverse Weather Adaptation, 3.1\% in Synthetic Source Adaptation, and approximately 2\% in both Distinct Camera and Diverse Context Adaptations. Detailed descriptions and evaluations of these four adaptation scenarios are provided in Section \ref{sec:experiments}. Moreover, comparing with methods using other detectors such as Faster R-CNN \cite{ren15faster}, Def DETR \cite{zhu21deformable}, and RetinaNet \cite{lin17focal}, DSBCO exhibits competitive performance on cross-domain situations, confirming its effective generalization capability. Our contributions are summarized as follows:
\begin{itemize}
    \item We propose the DSBCO framework, which extends the CST strategy to both classification and regression tasks, effectively breaking the performance bottleneck caused by unreliable pseudo-labels.
    
    \item We construct CST upon the MT paradigm to address the training instability associated with dense predictions in multiple object scenarios.
    
    \item We introduce a extra normalization mechanism to resolve exploding losses caused by the wide numerical range of regression coordinates. 
    
    \item Extensive experiments demonstrate that DSBCO achieves considerable performance on four standard cross-domain situations. Additionally, we provide the theoretical generalization bound derivation for the regression task.
\end{itemize}

%-------------------------------------------------------------------------
% Figure Insertion Code
%-------------------------------------------------------------------------
%-------------------------------------------------------------------------

%% file: sec/2_related_work.tex
\section{Related Work}
\label{sec:relate_work}

\noindent\textbf{Domain Adaptive Object Detection (DAOD)} aims to transfer learned visual representations from labeled source domains to unlabeled targets under challenging distribution shifts \cite{He23BiADT, Zhou26}. 
To relax these shifts, adversarial feature alignment matches feature distributions to extract and explicitly align shared domain-invariant representations from both domains \cite{Liu26, Feng2024DSDDA, Dong25Reconstructing}. 
In parallel, image-to-image translation synthesizes target-style images or disentangles domain-specific frequencies to harmonize the inputs at the pixel level \cite{qu25cfnet}. 
Beyond these global alignment strategies, graph-based methods utilize graph matching or relational learning to explicitly enforce cross-domain structural consistency, effectively aligning semantic topologies across domains \cite{xu2025lscdigm, Wang2025Optimal, Bi26}. 
Furthermore, transformer-based approaches leverage self-attention mechanisms to learn robust features that handle complex domain degradation \cite{li24react, Weng2024MTM, zeng2024acct, yao25sfod}. 
Among these diverse adaptation methods, ST effectively generates pseudo-labels on the target data to supervise the model iteratively, gradually narrowing the domain gap through reliable high-confidence predictions \cite{yoon24enhancing, Wang2025Deep, chen2025mgcamt, Amini2025Self}. 
Inspired by this success, our proposed method is primarily built upon the foundation of ST.

\noindent\textbf{Self-Training (ST)} serves as an effective paradigm in semi-supervised learning and domain adaptation \cite{liu21cycle, Amini2025Self}. 
Early strategies typically employ a single model to generate pseudo-labels and iteratively refine the network by selecting high-confidence predictions on unlabeled target data \cite{Zheng26, Zhang26, Dong25Reconstructing}. 
To enhance the stability of this process, the MT framework emerges as a robust and effective approach by maintaining a teacher model through the EMA of student parameters \cite{Tarvainen18, Li2022AT, Cao2023CMT, cai21eman}. 
This temporal ensemble mechanism provides more consistent supervision signals and effectively handles the noise inherent in pseudo-labels \cite{chen2025mgcamt, He2022TDD, deng2023harmonious}. 
Recent advances in this field further improve the reliability of supervision by incorporating clustering filters to remove outliers or adopting decoupled training strategies to balance classification and localization tasks \cite{Zhang26b, Zhou23ssda}. 
However, directly applying these co-training or self-training methods to object detection often leads to severe training instability and potential loss explosion due to the wide numerical range of regression targets \cite{yoo2022oada, Weng2024MTM, yoon24enhancing, wang24adt}. To address the problem, our DSBCO framework builds upon the stable MT paradigm to ensure effective adaptation across both classification and  regression tasks.

%% file: sec/3_methodology.tex
\section{Methodology}
\label{sec:methodology}

\subsection{Preliminaries}

The DAOD task involves a labeled source domain $\mathcal{D}_s = \{(x_s^i, y_s^i)\}_{i=1}^{N_s}$, where $x_s^i$ represents the input image and $y_s^i = \{y_{s,i}^{cls}, y_{s,i}^{reg}\}$ contains the category labels and bounding box coordinates. Meanwhile, an unlabeled target domain $\mathcal{D}_t = \{x_t^j\}_{j=1}^{N_t}$ is available for adaptation. Both domains share the same object categories but differ in data distributions, denoted as $P(\mathcal{X}_s) \neq P(\mathcal{X}_t)$, due to the domain shift. 

\subsection{DSBCO Framework}
\label{sec:method}

In this section, we present the details of the proposed DSBCO framework. As illustrated in Fig.~\ref{fig:framework}, our framework follows a left-to-right processing pipeline and mainly consists of three components: 1)  Mean Teacher (MT) paradigm that provides stable pseudo-labels; 2)  Instance Consistency Distillation aimed at promoting sample-level prediction consistency; and 3)  Feature Bilevel-Cycle Optimization module for addressing distribution shifts at the feature level.

% ----------------------------------------------------------------
% Part 1: Foundation (FCOS + Mean Teacher)
% ----------------------------------------------------------------

\subsubsection{MT Paradigm}
As illustrated in Fig.~\ref{fig:motivation}~(b), directly applying CST to dense prediction object detectors leads to extreme training instability and even model collapse (\textbf{Challenge \ding{183}}). To address this instability problem caused by dense prediction, we adopt  MT paradigm. In MT paradigm, the parameters of the Teacher model are updated via the EMA of the Student model. The update rule is formally defined as:
\begin{equation}
    \theta_{tea}^{(k)} = \alpha \theta_{tea}^{(k-1)} + (1 - \alpha) \theta_{stu}^{(k)},
\end{equation}
where $\alpha$ represents the smoothing coefficient, and $\theta_{tea}^{(k)}$ and $\theta_{stu}^{(k)}$ denote the parameters of the Teacher and Student networks at $k$-th iteration. This temporal smoothing process ensures prediction consistency across iterations, thereby providing stable pseudo-label  for the target domain. Furthermore, compared to the structurally complex two-stage Faster R-CNN, the anchor-free FCOS detector directly decouples the  task into independent pixel-level classification and regression predictions in a fully convolutional method. This characteristic aligns well with our proposed dual-stream optimization strategy, so we select it as the base architecture.

In terms of the data flow direction, the Teacher model only receives weakly augmented (\textit{e.g.}, basic image resizing and horizontal flipping) target domain images $\mathcal{A}_w$ to generate reliable pseudo-labels. Meanwhile, the trainable Student model receives source domain images $x_s$ and target domain images $x_t$ processed by strong augmentation $\mathcal{A}_s$ (\textit{e.g.}, introducing severe perturbations like color jittering, Gaussian blur, and random erasing).

The Student model is initially trained on the strongly augmented source images in a supervised way. The supervised loss $\mathcal{L}_{sup}$ is the sum of the classification loss and the regression loss:
\begin{equation}
    \mathcal{L}_{sup} =  \mathcal{L}_{cls1} + \mathcal{L}_{reg1}.
\end{equation}
Specifically, the classification task employs the Focal Loss \cite{lin17focal}:
\begin{equation}
    \mathcal{L}_{cls1} = - (1 - p_s)^{\gamma} \log(p_s),
\end{equation}
where $p_s$ denotes the probability predicted by the Student, and $\gamma$ is the focusing hyperparameter. For the regression task, to overcome the scale sensitivity of the traditional $L_1$ loss and the  vanishing gradient problem of the standard IoU, we adopt the Generalized Intersection over Union (GIoU) loss \cite{rezatofighi19giou}:
\begin{equation}
    \mathcal{L}_{reg1} = 1 - \left( \frac{|B_s \cap B^*|}{|B_s \cup B^*|} - \frac{|C \setminus (B_s \cup B^*)|}{|C|} \right),
\end{equation}
where $B_s$ and $B^*$ denote the predicted bounding box and the ground-truth bounding box, respectively, and $C$ is the smallest enclosing box covering both. The GIoU provides global regression constraints and continuous gradients, which are crucial for the early training stability of anchor-free models.

% ----------------------------------------------------------------
% Part 2: Instance Consistency (Separated)
% ----------------------------------------------------------------

\subsubsection{Instance Consistency Distillation}

Building upon the stable MT paradigm, we employ an instance consistency distillation mechanism as the baseline for self-training. This mechanism aims to enforce consistency between the Student model's predictions under strong augmentation and the Teacher model's pseudo-labels at the instance level.

Given that the Teacher model may produce unreliable predictions containing substantial background noise on the target domain, we first generate a binary mask $\mathbb{M}$ using the Teacher model's predicted classification probability $\hat{p}_t$ and localization quality $\hat{c}_t$ (\textit{i.e.}, centerness) to retain only high-quality samples:
\begin{equation}
    \mathbb{M} = \mathbb{I}\left(( \hat{p}_t > \tau_{cls}) \land (\hat{c}_t > \tau_{reg}) \right),
\end{equation}
where $\mathbb{I}(\cdot)$ denotes the indicator function (evaluating to $1$ if the condition holds, retaining the sample)~\cite{wang2023consistent}, and $\tau_{cls}$ and $\tau_{reg}$ are the predefined thresholds for classification probability and localization confidence, respectively. The instance consistency loss $\mathcal{L}_{distill}$ is computed within the valid regions defined by $\mathbb{M}$:
\begin{equation}
    \mathcal{L}_{distill} = \frac{1}{\sum \mathbb{M}} \sum \mathbb{M} \cdot \left( \mathcal{L}_{cls2} + \mathcal{L}_{reg2} \right).
\end{equation}
For classification alignment loss $\mathcal{L}_{cls2}$, since the Teacher model  provides continuous soft labels, we directly minimize the discrepancy by computing  Quality Focal Loss between the Student model's prediction $p_s$ and the Teacher model's soft label $\hat{p}_t$\cite{li20gfl}. The formula is as follows:
\begin{equation}
    \mathcal{L}_{cls2} = - \left| \hat{p}_t - p_s \right|^\beta \left[ \hat{p}_t \log(p_s) + (1 - \hat{p}_t) \log(1 - p_s) \right],
\end{equation}
where $\beta$ is a modulating factor used to suppress the contribution of easy samples where the Student model's prediction is already highly consistent with the Teacher model's soft label \cite{gorgizadeh21bias}. For regression consistency loss $\mathcal{L}_{reg2}$, we similarly compute the GIoU loss between the Student model's predicted boxes and the Teacher model's pseudo-label boxes.

% Part 3: Core Method (DSBCO)
% ----------------------------------------------------------------
\subsubsection{Feature Bilevel-Cycle Optimization}
As shown in Fig.~\ref{fig:motivation}~(a), standard self-training exhibits a performance upper bound in object detection due to its heavy reliance on unreliable pseudo-labels (\textbf{Challenge \ding{182}}). To overcome this, we propose  Feature Bilevel-Cycle Optimization strategy. This method formulates domain adaptation as a bilevel optimization problem. Unlike the original image level classification method CST, we leverage the decoupling characteristics of FCOS to design a dual-stream strategy containing classification and regression tasks, achieving optimization in the feature space. 

\paragraph{Classification Cycle Stream.} 
This branch is responsible for aligning semantic category features. In the inner loop (Inner Ridge), we abandon the unstable sample kernel matrix inversion method used in traditional CST and instead construct a covariance matrix along the feature dimension. We aim to find an optimal linear projection matrix $\mathbf{W}_{cls} \in \mathbb{R}^{d\times C}$ that can reconstruct the Teacher's pseudo-label vectors $\hat{p}_t$ from the Student's target domain features $\mathbf{Z}_{t} \in \mathbb{R}^{N_{t}\times d}$. This is formulated as a ridge regression problem with a closed-form solution:
\begin{equation}
    \mathbf{W}_{cls} = ( (\mathbf{Z}_t)^\top \mathbf{Z}_t + \lambda I )^{-1} (\mathbf{Z}_t)^\top \hat{p}_t,
\end{equation}
where $\lambda$ is the regularization coefficient. 
In the outer loop (Outer Projection), we directly apply this explicit mapping head $\mathbf{W}_{cls}$ learned on target domain to the source domain features $\mathbf{Z}_s \in \mathbb{R}^{N_{s}\times d}$, and enforce the projected results to align with source ground-truth labels. The classification cycle loss is defined as:
\begin{equation}
    \mathcal{L}_{cstcls} = \| \mathbf{Z}_s \mathbf{W}_{cls} - y_s \|_F^2,
\end{equation}
where $\| \cdot \|_F^2$ denotes the squared Frobenius norm of a matrix ~\cite{wang2019few}. By minimizing this loss, we encourage the network to learn domain-invariant semantic features.

\paragraph{Regression Cycle Stream.} 
As shown in Fig.~\ref{fig:motivation}~(c), due to the unbounded nature and wide numerical range of bounding box regression offsets, directly applying cycle consistency to the regression task leads to severe exploding losses (\textbf{Challenge \ding{184}}). To address this issue, we introduce a regression normalization mechanism in this stream. In the inner loop, we first compute the mean $\mu_t$ and standard deviation $\sigma_t$ of the target domain regression offsets $\hat{\delta}_t$ generated by the Teacher model, and normalize them into a standard normal distribution:
\begin{equation}
    \tilde{\delta}_t = \frac{\hat{\delta}_t - \mu_t}{\sigma_t + \epsilon},
\end{equation}
where $\epsilon$ is a small constant to prevent division by zero. Subsequently, we utilize these normalized targets to solve for the optimal regression matrix $\mathbf{W}_{reg} \in \mathbb{R}^{d\times 4}$:
\begin{equation}
\mathbf{W}_{reg} = ( (\mathbf{Z}_t)^\top \mathbf{Z}_t + \lambda I )^{-1} (\mathbf{Z}_t)^\top \tilde{\delta}_t.
\end{equation}
During the outer loop projection, we normalize the source domain's ground-truth labels $y_s$ using the same target domain statistics ($\mu_t, \sigma_t$) to obtain $\tilde{y}_s$. Finally, the regression cycle loss $\mathcal{L}_{cstreg}$ is computed as:
\begin{equation}
    \mathcal{L}_{cstreg} = \| \mathbf{Z}_s \mathbf{W}_{reg} - \tilde{y}_s \|_F^2.
\end{equation}
Through this normalization and reconstruction mechanism, we map the large regression values into a stable space, effectively preventing gradient divergence. This effectively extends CST to the localization task in object detection.

%----------------------------------------------------------------
% Part 4: Total Objective & Algorithm
% ----------------------------------------------------------------
\subsubsection{Total Optimization Objective}
The overall optimization objective $\mathcal{L}_{total}$ combines the supervised loss from the source domain, the instance level consistency distillation loss, and the feature level cycle alignment loss. The complete loss function is formulated as follows:
\begin{equation}
\label{eq:final_total}
\begin{split}
    \mathcal{L}_{total} = \mathcal{L}_{sup} &+ \lambda_{unsup} \mathcal{L}_{distill} + \lambda_{cycle} \big( \mathcal{L}_{cstcls} + \eta \mathcal{L}_{cstreg} \big),
\end{split}
\end{equation}
where $\lambda_{unsup}$ and $\lambda_{cycle}$ are employed to balance the contributions of the unsupervised signals. Meanwhile, $\eta$ is a hyperparameter specifically designed to adjust the relative importance of the regression cycle alignment. The complete training process is detailed in Algorithm~\ref{alg:training_process}.

\begin{algorithm}[t]

\caption{DSBCO Algorithm}
\label{alg:training_process}

\begin{algorithmic}[1]

\renewcommand{\algorithmicrequire}{\textbf{Input:}}
\renewcommand{\algorithmicensure}{\textbf{Output:}}

\REQUIRE Labeled source domain $\mathcal{D}_s = \{(x_s, y_s)\}$, Unlabeled target domain $\mathcal{D}_t = \{x_t\}$, Max Iterations $T_{max}$, Burn-in $T_{burn}$, learning rate $l_r$
\ENSURE Optimized Student Parameters $\theta_{stu}^*$.
\STATE $\theta_{stu} \leftarrow \text{RandomInit}(.)$, $\theta_{tea} \leftarrow \theta_{stu}$. // \textbf{Initialization}

\FOR{$t = 1$ \textbf{to} $T_{max}$}
    \STATE $\{(x_s, y_s)\} \sim \mathcal{D}_s$ \& $\{x_t\} \sim \mathcal{D}_t$ // \textbf{Sample Batch}
    
    \item[] \hspace{-2.2em} \colorbox{gray!25}{\makebox[\dimexpr\linewidth+2.2em][l]{\textbf{/* Instance Consistency Distillation */}}}
    
    \STATE $\hat{p}_t, \hat{\delta}_t \leftarrow F_{\theta_{tea}}(\mathcal{A}_w(x_t))$ 
    \STATE $\mathbb{M} = \mathbb{I}\left(( \hat{p}_t  > \tau_{cls}) \land (\hat{c}_t > \tau_{reg}) \right)$ 
    \STATE $\mathcal{L}_{sup} \leftarrow (F_{\theta_{stu}}(\mathcal{A}_s(x_s)), y_s)$ 
    \STATE $\mathcal{L}_{distill} \leftarrow (\hat{p}_t, \hat{\delta}_t, F_{\theta_{stu}}(\mathcal{A}_s(x_t)), \mathbb{M})$ 
    \STATE $\mathcal{L}_{total} = \mathcal{L}_{sup} + \lambda_{unsup} \mathcal{L}_{distill}$

    \item[] \hspace{-2.2em} \colorbox{gray!25}{\makebox[\dimexpr\linewidth+2.2em][l]{\textbf{/* Distribution Bilevel-Cycle Optimization */}}}
    
    \IF{$t > T_{burn}$}
        \STATE  $\mathbf{Z}_s \leftarrow \Phi_{\theta_{stu}}(x_s)$ \& $\mathbf{Z}_t \leftarrow \Phi_{\theta_{stu}}(x_t)$ // \textbf{Extract Features}
        
        // \textbf{Classification Cycle Stream}
        \STATE $\mathbf{W}_{cls} = ((\mathbf{Z}_t)^\top \mathbf{Z}_t + \lambda \mathbf{I})^{-1} (\mathbf{Z}_t)^\top \hat{p}_t$ 
        \STATE $\mathcal{L}_{cstcls} = \| \mathbf{Z}_s \mathbf{W}_{cls} - y_s \|_F^2$ 
        
        // \textbf{Regression Cycle Stream}
        \STATE $\tilde{\delta}_t = (\hat{\delta}_t - \mu_t) / (\sigma_t + \epsilon)$  // \textbf{Normalization}
        \STATE $\mathbf{W}_{reg} = ((\mathbf{Z}_t)^\top \mathbf{Z}_t + \lambda \mathbf{I})^{-1} (\mathbf{Z}_t)^\top \tilde{\delta}_t$ 
        \STATE $\hat{y}_s = (y_s - \mu_t) / (\sigma_t + \epsilon)$ // \textbf{Normalization}
        \STATE $\mathcal{L}_{cstreg} = \| \mathbf{Z}_s \mathbf{W}_{reg} - \hat{y}_s \|_F^2$ 
        
        \STATE $\mathcal{L}_{total} = \mathcal{L}_{total} + \lambda_{cycle}(\mathcal{L}_{cstcls} + \eta \mathcal{L}_{cstreg})$
    \ENDIF

    \item[] \hspace{-2.2em} \colorbox{gray!25}{\makebox[\dimexpr\linewidth+2.2em][l]{\textbf{/* Parameter Optimization */}}}
    
    \STATE $\theta_{stu} = \theta_{stu} - l_r \nabla \mathcal{L}_{total}$
    \STATE $\theta_{tea} = \alpha \theta_{tea} + (1-\alpha) \theta_{stu}$
\ENDFOR

\STATE \textbf{return} $\theta_{stu}^*$

\end{algorithmic}
\end{algorithm}

\subsection{Theoretical Regression Bound}
\label{sec:theory}

Extending generalization bounds to continuous bounding box regression is non-trivial due to the infinite output space and optimization instability. We therefore analyze DSBCO by bounding the target regression error, characterizing the failure of standard ST, and showing how cycle optimization tightens the bound.

\begin{theorem}[Generalization Upper Bound]
Under the Lipschitz continuity assumption for continuous box regression, the expected target error $\mathcal{E}_{\mathcal{Q}}(f_s)$ is bounded by the sum of cycle reconstruction error and stability terms:

\begin{equation}
    \mathcal{E}_{\mathcal{Q}}(f_s) \le 
    \hat{\mathcal{E}}_{\mathcal{P}}(f_t) +
    \hat{c}_{\mathcal{Q}}(f_s, f_t) +
    \mathcal{R}(f_t,\delta) +
    \mathcal{O}(\mathfrak{R}_n(\mathcal{F})).
\end{equation}
Here, $\mathcal{E}_{\mathcal{Q}}(f_s)=\mathbb{E}_{(x,y)\sim\mathcal{Q}}[\mathcal{L}_{reg}(f_s(x),y)]$ is the target regression error, $\hat{\mathcal{E}}_{\mathcal{P}}(f_t)=\frac{1}{n_s}\sum_i\mathcal{L}_{reg}(f_t(x_s^i),y_s^i)$ is the empirical source cycle reconstruction error, $\hat{c}_{\mathcal{Q}}(f_s,f_t)=\frac{1}{n_t}\sum_j d_{box}(f_s(x_t^j),f_t(x_t^j))$ measures the empirical student-teacher discrepancy on target samples, $\mathcal{R}(f_t,\delta)$ denotes the probability of large box shifts under local perturbations, and $\mathfrak{R}_n(\mathcal{F})$ is the Rademacher complexity of the detector class.
\end{theorem}

\begin{theorem}[Failure of Standard ST]
Under a hard scenario where target pseudo-labels are dominated by background correlations, standard ST converges to a spurious solution $f_{spurious}$ with an irreducible target regression error:
\begin{equation}
    \mathcal{E}_{\mathcal{Q}}(f_{ST}) \ge \epsilon > 0.
\end{equation}
\end{theorem}

\begin{theorem}[Convergence of DSBCO]
Minimizing the cycle consistency term $\hat{\mathcal{E}}_{\mathcal{P}}(f_t)$ acts as a source-domain verification mechanism. A spurious solution incurs high source reconstruction error, whereas the causal object-semantic solution preserves localization semantics and approaches zero cycle loss.
\end{theorem}

\begin{theorem}[Comprehensive Generalization Bound for DSBCO]
By combining the regression generalization bound (Theorem 1) and the robustness analysis, we prove that optimizing the DSBCO objective $\mathcal{L}_{\text{DSBCO}}$ directly minimizes the dominant empirical terms in the target error upper bound:

\begin{equation}
    \mathcal{E}_{\mathcal{Q}}(f_s) \le 
    \mathcal{L}_{\text{DSBCO}} +
    \mathcal{O}(\mathfrak{R}_n(\mathcal{F})) + \text{const}.
\end{equation}
\end{theorem}

\noindent\textbf{Theoretical Analysis.}
Complete proofs are provided in \textbf{Section A} of the supplementary material. \textbf{Theorem 1} formally shows that target localization error is controlled by source cycle reconstruction, student-teacher discrepancy, and regression non-robustness. \textbf{Theorem 2} further explains why standard ST can be trapped by background-biased pseudo-labels. Furthermore, \textbf{Theorem 3 and Theorem 4} show that DSBCO tightens the bound through complementary mechanisms. Specifically, \textbf{Theorem 3} uses cycle losses to reject spurious mappings and stabilize source-target regression correspondence, while \textbf{Theorem 4} employs instance consistency distillation to reduce $\hat{c}_{\mathcal{Q}}$ and dual-stream thresholding to effectively suppress $\mathcal{R}(f_t,\delta)$ against noisy pseudo-label perturbations.

%% file: sec/4_experiments.tex
\section{Experiments}
\label{sec:experiments}

In this section, we present a comprehensive evaluation of the proposed DSBCO framework. We evaluate DSBCO on four different domain adaptation scenarios to test the model's robustness against weather changes, scene variations, camera differences, and synthetic-to-real shifts. 

%------------------------------------------------------------------------
% Table: Cityscapes -> Foggy Cityscapes
%------------------------------------------------------------------------
\begin{table}[t] 
\centering
\caption{Quantitative results on \textbf{Adverse Weather Adaptation} (Cityscapes $\to$ Foggy Cityscapes). The mean Average Precision (\%) is evaluated on the target domain. The best and second-best results are highlighted in \textbf{bold} and \underline{underlined}, respectively.}
\label{tab:foggy}

\setlength{\tabcolsep}{3.2pt} 
\resizebox{\linewidth}{!}{%
\begin{tabular}{lcc|cccccccc|c}
\specialrule{1.5pt}{1pt}{1pt} 

\textbf{Method} & \textbf{Venue} & \textbf{Detector} & \textbf{Person} & \textbf{Rider} & \textbf{Car} & \textbf{Truck} & \textbf{Bus} & \textbf{Train} & \textbf{Motor} & \textbf{Bicycle} & \textbf{mAP} $\uparrow$ \\ 
\specialrule{1.3pt}{1pt}{1pt}

D\_adapt \cite{Jiang2022Dadapt} & ICLR '22 & Faster R-CNN & 44.9 & 54.2 & 61.7 & 25.6 & 36.3 & 24.7 & 37.3 & 46.1 & 41.3 \\
TDD \cite{He2022TDD} & CVPR '22 & Faster R-CNN & 39.6 & 47.5 & 55.7 & 33.8 & 47.6 & 42.1 & 37.0 & 41.4 & 43.1 \\
AT \cite{Li2022AT} & CVPR '22 & Faster R-CNN & 45.5 & 55.1 & 64.2 & 35.0 & 56.3 & 54.3 & 38.5 & 51.9 & 50.9 \\
PT \cite{Chen2022PT} & ICML '22 & Faster R-CNN & 40.2 & 48.8 & 59.7 & 30.7 & 51.8 & 30.6 & 35.4 & 44.5 & 42.7 \\
CMT \cite{Cao2023CMT} & CVPR '23 & Faster R-CNN & 45.9 & 55.7 & 63.7 & \underline{39.6} & \underline{66.0} & 38.8 & 41.4 & 51.2 & 50.3 \\
AT + DSD-DA \cite{Feng2024DSDDA} & ICML '24 & Faster R-CNN & 49.1 & 59.3 & 66.2 & 35.8 & 60.0 & 47.1 & 45.2 & 54.9 & 52.2 \\
CMT + DSD-DA \cite{Feng2024DSDDA} & ICML '24 & Faster R-CNN & 49.0 & 59.6 & 65.3 & 35.7 & 61.0 & 46.5 & 43.9 & 57.3 & 52.3 \\
AT + REACT \cite{li24react} & TIP '24 & Faster R-CNN & 51.4 & 57.9 & 67.4 & 37.7 & 58.4 & 52.8 & 44.6 & 54.6 & 53.1 \\
\midrule
AQT \cite{Huang2022AQT} & IJCAI '22 & Def DETR & 49.3 & 52.3 & 64.4 & 27.7 & 53.7 & 46.5 & 36.0 & 46.4 & 47.1 \\
O$^2$Net \cite{Gong2022O2Net} & ACM MM '22 & Def DETR & 48.7 & 51.5 & 63.6 & 31.1 & 47.6 & 47.8 & 38.0 & 45.9 & 46.8 \\
MTTrans \cite{yu2022mttrans} & ECCV '22 & Def DETR & 47.7 & 49.9 & 65.2 & 25.8 & 45.9 & 33.9 & 32.6 & 43.5 & 46.8 \\
DA-DETR \cite{Zhang2023DADETR} & CVPR '23 & Def DETR & 49.9 & 50.0 & 63.1 & 24.0 & 45.8 & 37.5 & 31.6 & 46.3 & 43.5 \\
BIADT \cite{He23BiADT} & ICCV '23 & Def DETR & 52.2 & 58.9 & 69.2 & 31.7 & 55.0 & 45.1 & 42.6 & 51.3 & 50.8 \\
MRT \cite{Zhao2023MRT} & ICCV '23 & Def DETR & 52.8 & 51.7 & 68.7 & 35.9 & 58.1 & 54.5 & 41.0 & 47.1 & 51.2 \\
MTM \cite{Weng2024MTM} & AAAI '24 & Def DETR & 51.0 & 53.4 & 67.2 & 37.2 & 54.4 & 41.6 & 38.4 & 47.7 & 48.9 \\
ACCT \cite{zeng2024acct} & ICASSP '24 & Def DETR & 53.6 & 58.9 & 69.4 & 31.1 & 53.5 & 33.7 & 42.6 & 54.4 & 49.6 \\
\midrule
KTNet\cite{tian2023ktnet} & Neural Networks '23 & RetinaNet & 43.7 & 48.9 & 56.9 & 34.9 & 51.8 & \underline{55.3} & 32.8 & 42.8 & 45.9 \\
MGCAMT \cite{chen2025mgcamt} & TIP '25 & RetinaNet & \underline{60.2} & \underline{66.6} & \textbf{76.5} & 33.2 & 60.1 & 43.2 & \underline{49.8} & \underline{57.9} & \underline{55.9} \\
\midrule
Source Only & - & FCOS & 38.5 & 41.8 & 46.0 & 28.6 & 42.1 & 11.0 & 31.1 & 42.2 & 35.1 \\
SIGMA \cite{li2022sigma} & CVPR '22 & FCOS & 44.0 & 43.9 & 60.3 & 31.6 & 50.4 & 51.5 & 31.7 & 40.6 & 44.2 \\
OADA \cite{yoo2022oada} & ECCV '22 & FCOS & 47.8 & 46.5 & 62.9 & 32.1 & 48.5 & 50.9 & 34.3 & 39.8 & 45.4 \\
SIGMA++ \cite{li2023sigma++} & TPAMI '23 & FCOS & 46.4 & 45.1 & 61.0 & 32.1 & 52.2 & 44.6 & 34.8 & 39.9 & 44.5 \\
CIGAR \cite{liu2023cigar} & CVPR '23 & FCOS & 46.1 & 47.3 & 62.1 & 27.8 & 56.6 & 44.3 & 33.7 & 41.3 & 44.9 \\
HT \cite{deng2023harmonious} & CVPR '23 & FCOS & 52.1 & 55.8 & 67.5 & 32.7 & 55.9 & 49.1 & 40.1 & 50.3 & 50.4 \\
LSCDIGM \cite{xu2025lscdigm} & J SUPERCOMPUT '25 & FCOS & 44.7 & 43.1 & 60.9 & 28.9 & 48.3 & 48.0 & 36.7 & 37.7 & 43.5 \\
DSTS \cite{Dong25Reconstructing} & Information '25 & FCOS & 51.4 & 54.3 & 68.4 & 37.8 & 54.6 & 40.2 & 42.4 & 51.8 & 50.1 \\
\rowcolor{gray!20} \textbf{DSBCO (Ours)} & - & FCOS & \textbf{61.1} & \textbf{66.8} & \underline{73.0} & \textbf{58.1} & \textbf{72.3} & \textbf{62.2} & \textbf{58.8} & \textbf{65.6} & \textbf{64.7} \\
\specialrule{1.5pt}{1pt}{1pt}
\end{tabular}%
}
\end{table}
%----------------------------------------------------------
% Table: Cityscapes -> BDD100K 
%------------------------------------------------------------------------
\begin{table}[t] 
\centering
\caption{Quantitative results on \textbf{Diverse Context Adaptation} (Cityscapes $\to$ BDD100K). The mean Average Precision (\%) is evaluated on the target domain. The best and second-best results are highlighted in \textbf{bold} and \underline{underlined}, respectively.}
\label{tab:diverse_context}

\resizebox{0.95\linewidth}{!}{% 
\begin{tabular}{lcc|ccccccc|c}
\specialrule{1.5pt}{1pt}{1pt} 

\textbf{Method} & \textbf{Venue} & \textbf{Detector} & \textbf{Person} & \textbf{Rider} & \textbf{Car} & \textbf{Truck} & \textbf{Bus} & \textbf{Motor} & \textbf{Bicycle} & \textbf{mAP} $\uparrow$ \\ 
\specialrule{1.3pt}{1pt}{1pt}
TDD \cite{He2022TDD} & CVPR '22 & Faster R-CNN & 39.6 & 38.9 & 53.9 & 24.1 & 25.5 & 24.5 & 28.8 & 33.6 \\
PT \cite{Chen2022PT} & ICML '22 & Faster R-CNN & 40.5 & 39.9 & 52.7 & 25.8 & \underline{33.8} & 23.0 & 28.8 & 34.9 \\
AT + REACT \cite{li24react} & TIP '24 & Faster R-CNN & - & - & - & - & - & - & - & 35.8 \\
\midrule
AQT \cite{Huang2022AQT} & IJCAI '22 & Def DETR & 38.2 & 33.0 & 58.4 & 17.3 & 18.4 & 16.9 & 23.5 & 29.4 \\
O$^2$Net \cite{Gong2022O2Net} & ACM MM '22 & Def DETR & 40.4 & 31.2 & 58.6 & 20.4 & 25.0 & 14.9 & 22.7 & 30.5 \\
MTTrans \cite{yu2022mttrans} & ECCV '22 & Def DETR & 44.1 & 30.1 & 61.5 & 25.1 & 26.9 & 23.0 & 17.7 & 32.6 \\
BIADT \cite{He23BiADT} & ICCV '23 & Def DETR & 42.1 & 34.0 & 60.9 & 17.4 & 19.5 & 18.2 & 25.7 & 33.6 \\
MRT \cite{Zhao2023MRT} & ICCV '23 & Def DETR & 48.4 & 30.9 & 63.7 & 24.7 & 25.5 & 20.2 & 22.6 & 33.7 \\
MTM \cite{Weng2024MTM} & AAAI '24 & Def DETR & 53.7 & 35.1 & \underline{68.8} & 23.0 & 28.8 & 23.8 & 28.0 & 37.3 \\
ACCT \cite{zeng2024acct} & ICASSP '24 & Def DETR & 51.8 & 41.4 & 61.8 & 26.0 & 23.4 & \textbf{36.9} & 31.7 & 39.0 \\
\midrule
MGCAMT \cite{chen2025mgcamt} & TIP '25 & RetinaNet & \textbf{61.2} & \textbf{44.9} & \textbf{75.3} & \textbf{32.5} & 30.0 & \underline{31.9} & \underline{37.8} & \textbf{44.8} \\
\midrule
Source Only & - & FCOS & 36.9 & 22.4 & 49.7 & 16.1 & 16.3 & 13.0 & 22.1 & 25.2 \\
SIGMA \cite{li2022sigma} & CVPR '22 & FCOS & 46.9 & 29.6 & 64.1 & 20.2 & 23.6 & 17.9 & 26.3 & 32.7 \\
HT \cite{deng2023harmonious} & CVPR '23 & FCOS & 53.4 & 40.4 & 63.5 & 27.4 & 30.6 & 28.2 & \textbf{38.0} & 40.2 \\
SIGMA++ \cite{li2023sigma++} & TPAMI '23 & FCOS & 47.5 & 30.4 & 65.6 & 21.1 & 26.3 & 27.1 & 17.8 & 33.7 \\
\rowcolor{gray!25} \textbf{DSBCO (Ours)} & - & FCOS & 54.8 & 38.2 & 67.9 & \underline{32.3} & \textbf{35.1} & 29.3 & 35.7 & \underline{41.9} \\
\specialrule{1.5pt}{1pt}{1pt}
\end{tabular}%
}
\end{table}

\begin{table}[t]
    \centering

    % --- Table 3 (Left) ---
    \begin{minipage}[t]{0.47\linewidth}
        \centering
        \caption{Quantitative results on \textbf{Distinct Camera Adaptation} (KITTI $\to$ Cityscapes). Average Precision (\%) is evaluated on the target domain. The best and second-best results are highlighted in \textbf{bold} and \underline{underlined}, respectively.}
        \label{tab:distinct_camera}
        \begingroup
        \renewcommand{\arraystretch}{1.1}
        \resizebox{\linewidth}{!}{%
            \begin{tabular}{lcc|c}
            \specialrule{1.5pt}{1pt}{1pt}
            \textbf{Method} & \textbf{Venue} & \textbf{Detector} & \textbf{AP of Car} $\uparrow$ \\ 
            \specialrule{1.3pt}{1pt}{1pt}
            TDD \cite{He2022TDD} & CVPR '22 & Faster R-CNN & 47.4 \\
            PT \cite{Chen2022PT} & ICML '22 & Faster R-CNN & 60.2 \\
            AT + DSD-DA \cite{Feng2024DSDDA} & ICML '24 & Faster R-CNN & 49.3 \\
            AT + REACT \cite{li24react} & TIP '24 & Faster R-CNN & 59.5 \\
            \hline
            DA-DETR \cite{Zhang2023DADETR} & CVPR '23 & Def DETR & 48.9 \\
            \hline
            MGCAMT \cite{chen2025mgcamt} & TIP '25 & RetinaNet & \textbf{62.2} \\
            \hline
            Source Only & - & FCOS & 34.4 \\
            SIGMA \cite{li2022sigma} & CVPR '22 & FCOS & 45.8 \\
            OADA \cite{yoo2022oada} & ECCV '22 & FCOS & 47.8 \\
            CIGAR \cite{liu2023cigar} & CVPR '23 & FCOS & 48.5 \\
            HT \cite{deng2023harmonious} & CVPR '23 & FCOS & \underline{60.3} \\
            LSCDIGM \cite{xu2025lscdigm} & J SUPERCOMPUT '25 & FCOS & 50.0 \\
            DSTS \cite{Dong25Reconstructing} & Information '25 & FCOS & 50.9 \\
            \rowcolor{gray!25} \textbf{DSBCO (Ours)} & - & FCOS & \textbf{62.2} \\
            \specialrule{1.5pt}{1pt}{1pt}
            \end{tabular}
        }
        \endgroup
    \end{minipage}
    \hfill
    % --- Table 4 (Right) ---
    \begin{minipage}[t]{0.50\linewidth}
        \centering
        \caption{Quantitative results on \textbf{Synthetic Source Adaptation} (Sim10K $\to$ Cityscapes). Average Precision (\%) is evaluated on the target domain. The best and second-best results are highlighted in \textbf{bold} and \underline{underlined}, respectively.}
        \label{tab:synthetic_source}
        \begingroup
        \renewcommand{\arraystretch}{1.00}
        \resizebox{\linewidth}{!}{%
            \begin{tabular}{lcc|c}
            \specialrule{1.5pt}{1pt}{1pt}
            \textbf{Method} & \textbf{Venue} & \textbf{Detector} & \textbf{AP of Car} $\uparrow$ \\ 
            \specialrule{1.3pt}{1pt}{1pt}
            D\_adapt \cite{Jiang2022Dadapt} & ICLR '22 & Faster R-CNN & 50.3 \\
            TDD \cite{He2022TDD} & CVPR '22 & Faster R-CNN & 53.4 \\
            PT \cite{Chen2022PT} & ICML '22 & Faster R-CNN & 55.1 \\
            AT + DSD-DA \cite{Feng2024DSDDA} & ICML '24 & Faster R-CNN & 52.5 \\
            MT + DAS \cite{yu2024das} & NeurIPS '24 & Faster R-CNN & 55.4 \\
            AT + REACT \cite{li24react} & TIP '24 & Faster R-CNN & 58.6 \\
            \hline
            O$^2$Net \cite{Gong2022O2Net} & ACM MM '22 & Def DETR & 54.1 \\
            DA-DETR \cite{Zhang2023DADETR} & CVPR '23 & Def DETR & 54.7 \\
            BIADT \cite{He23BiADT} & ICCV '23 & Def DETR & 56.6 \\
            MRT \cite{Zhao2023MRT} & ICCV '23 & Def DETR & 62.0 \\
            MTM \cite{Weng2024MTM} & AAAI '24 & Def DETR & 58.1 \\
            \hline
            MGCAMT \cite{chen2025mgcamt} & TIP '25 & RetinaNet & \underline{67.5} \\
            \hline
            Source Only & - & FCOS & 39.8 \\
            SIGMA \cite{li2022sigma} & CVPR '22 & FCOS & 53.7 \\
            OADA \cite{yoo2022oada} & ECCV '22 & FCOS & 59.2 \\
            CIGAR \cite{liu2023cigar} & CVPR '23 & FCOS & 58.5 \\
            HT \cite{deng2023harmonious} & CVPR '23 & FCOS & 65.5 \\
            LSCDIGM \cite{xu2025lscdigm} & J SUPERCOMPUT '25 & FCOS & 56.2 \\
            DSTS \cite{Dong25Reconstructing} & Information '25 & FCOS & 60.4 \\
            \rowcolor{gray!25} \textbf{DSBCO (Ours)} & - & FCOS & \textbf{68.6} \\
            \specialrule{1.5pt}{1pt}{1pt}
            \end{tabular}
        }
        \endgroup
    \end{minipage}

    \par\medskip

    % ===================== Row 2: Table 5 and Table 6 =====================

    % --- Table 5: Complexity ---
    \begin{minipage}[t]{0.45\linewidth}
        \centering
        \caption{Comparison of computational complexity and training time cost. Naive Bilevel refers to standard bilevel optimization, requiring $M$ internal gradient steps (typically $M \gg 1$, \textit{e.g.}, $M \approx 50$) per outer update. DSBCO bypasses these iterative inner loops for high efficiency. Here, $N$ is the number of samples, and $T$ is the unit training time of Standard ST.}
        \label{tab:time_cost}
        \resizebox{\linewidth}{!}{%
            \begin{tabular}{lcc}
            \specialrule{1.5pt}{1pt}{1pt}
            \textbf{Optimization Strategy} & \textbf{Complexity} & \textbf{Time Cost} \\ 
            \specialrule{1.3pt}{1pt}{1pt}
            Standard ST & $\mathcal{O}(N)$ & $T$ \\
            Naive Bilevel Optimization & $\mathcal{O}(N \cdot M)$ & $\approx M \cdot T$ \\
            \rowcolor{gray!25} \textbf{DSBCO (Ours)} & $\mathcal{O}(N)$ & \textbf{$\approx 1.2 \cdot T$} \\
            \specialrule{1.5pt}{1pt}{1pt}
            \end{tabular}
        }
    \end{minipage}
    \hfill
    % --- Table 6: Ablation ---
    \begin{minipage}[t]{0.52\linewidth}
        \centering
        \caption{Ablation study on \textbf{Adverse Weather Adaptation} (Cityscapes $\to$ Foggy Cityscapes). This table provides a comprehensive and incremental evaluation to analyze the individual contribution of Instance Consistency ($\mathcal{L}_{distill}$), the Classification Cycle ($\mathcal{L}_{cstcls}$), and the Regression Cycle ($\mathcal{L}_{cstreg}$). The mean Average Precision (\%) (mAP) is evaluated on the target domain.}
        \label{tab:ablation}
        \resizebox{\linewidth}{!}{%
            \begin{tabular}{lcccc|c}
            \specialrule{1.5pt}{1pt}{1pt}
            \textbf{Method} & \textbf{Baseline} & \textbf{$\mathcal{L}_{distill}$} & \textbf{$\mathcal{L}_{cstcls}$} & \textbf{$\mathcal{L}_{cstreg}$} & \textbf{mAP $\uparrow$} \\
            \specialrule{1.3pt}{1pt}{1pt}
            Standard ST & \checkmark & & & & 41.2 \\
            w/ Instance Consistency & \checkmark & \checkmark & & & 48.3 \\
            w/ Classification Cycle & \checkmark & \checkmark & \checkmark & & 54.5 \\
            w/ Regression Cycle & \checkmark & \checkmark & & \checkmark & 50.9 \\
            \rowcolor{gray!25} \textbf{DSBCO (Ours)} & \checkmark & \checkmark & \checkmark & \checkmark & \textbf{64.7} \\
            \specialrule{1.5pt}{1pt}{1pt}
            \end{tabular}
        }
    \end{minipage}
\end{table}

\subsection{Experimental Setup}

\noindent\textbf{Cross Domain Scenarios and Datasets.} We evaluate our framework on four standard benchmarks covering diverse domain shifts. 
\textbf{Adverse Weather Adaptation} transfers from \textbf{Cityscapes}~\cite{cordts16cityscapes}, which captures clear street scenes from 50 German cities, to \textbf{Foggy Cityscapes}~\cite{sakaridis18foggy}, synthesized with realistic fog density $\beta=0.02$, to test robustness against visibility degradation. 
\textbf{Diverse Context Adaptation} adapts Cityscapes to the daytime driving subset of \textbf{BDD100K}~\cite{yu20bdd100k}, collected across various locations with complex lighting conditions, addressing variations in geography and scene layout. 
\textbf{Distinct Camera Adaptation} utilizes \textbf{KITTI}~\cite{geiger12kitti}, characterized by a wide aspect ratio and unique sensor setup, as the source for Cityscapes to handle significant cross-camera geometric mismatches from differing camera intrinsics. 
Finally, \textbf{Synthetic Source Adaptation} leverages \textbf{Sim10K}~\cite{johnson17sim10k}, rendered from the game GTA V, to adapt to the target \textbf{Cityscapes} domain, bridging the synthetic-to-real texture gap. Details are provided in Supplementary \textbf{Section B}.

\noindent\textbf{Evaluation Metrics.}
Following standard protocols in cross-domain object detection, we employ the mean Average Precision (mAP) as the primary evaluation metric with an Intersection-over-Union (IoU) threshold of 0.5. For experiments involving the Cityscapes, Foggy Cityscapes, and BDD100K datasets, we report the mAP across all shared semantic categories. In the Synthetic Source Adaptation and Distinct Camera Adaptation scenarios, we report the Average Precision (AP) for the common category Car. This rigorous evaluation protocol ensures a fair and direct comparison with prior methodologies.

\noindent\textbf{Implementation Details and ST Baseline.} 
We implement FCOS with a VGG-16 backbone \cite{simonyan15vgg}, optimized via SGD with a learning rate of 0.004. Training consists of 30k pre-training iterations on the source data $\mathcal{D}_s$, followed by 90k adaptation iterations on a single RTX 3090 GPU. For the ST baseline, the pre-trained detector generates pseudo-labels for target data $\mathcal{D}_t$, filtering predictions with confidence scores below 0.45. The model is then iteratively retrained on both the labeled source and pseudo-labeled target data. Further implementation details are provided in   Supplementary \textbf{Section C}.

\subsection{Main Results}

\begin{figure*}[t]
    \centering
    \includegraphics[width=1.0\linewidth, height=0.4\textheight, keepaspectratio]{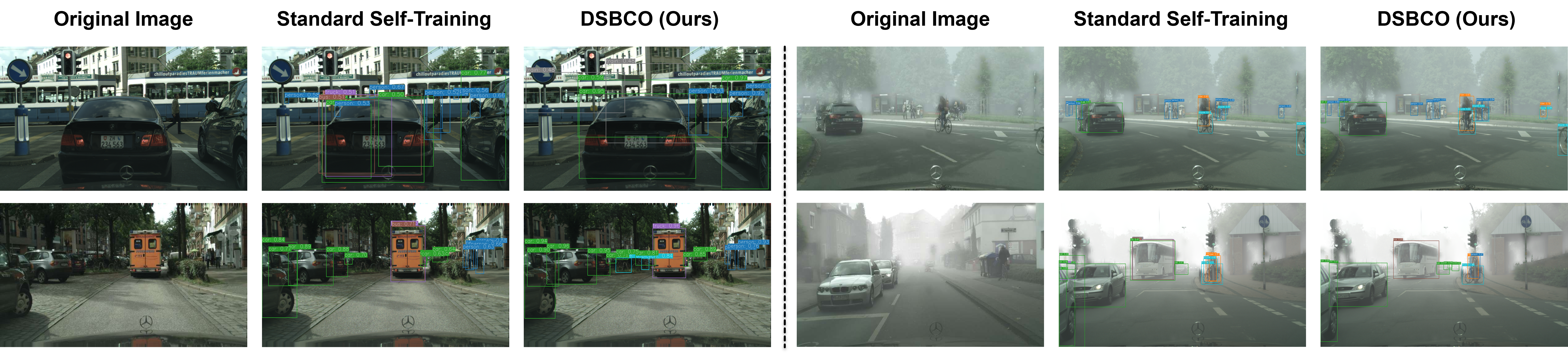} 
    \caption{Qualitative comparative detection results. DSBCO shows effective and robust improvement in resolving severe bounding box shifts and misclassifications.}
    \label{fig:vis_results}
\end{figure*}

\begin{figure}[t]
    \centering

    % ================= First Row Only =================
    \begin{minipage}[t]{0.48\linewidth}
        \centering
        \includegraphics[width=0.92\linewidth]{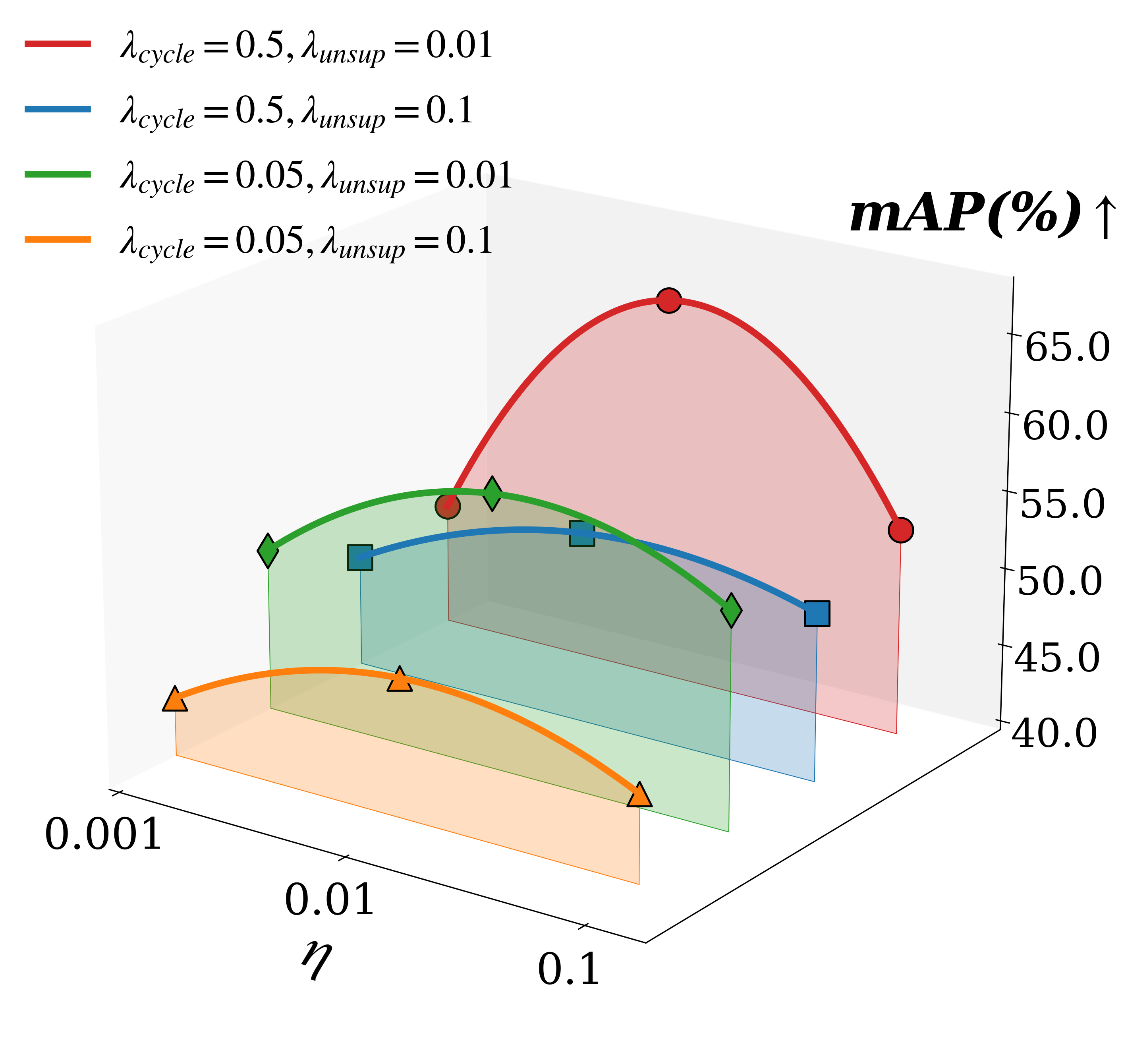}
        \captionsetup{skip=1pt}
        \captionof{figure}{Hyperparameter sensitivity analysis of our proposed DSBCO.}
        \label{fig:hyperparam_sensitivity}
    \end{minipage}
    \hfill
    \begin{minipage}[t]{0.48\linewidth}
        \centering
        \includegraphics[width=0.96\linewidth]{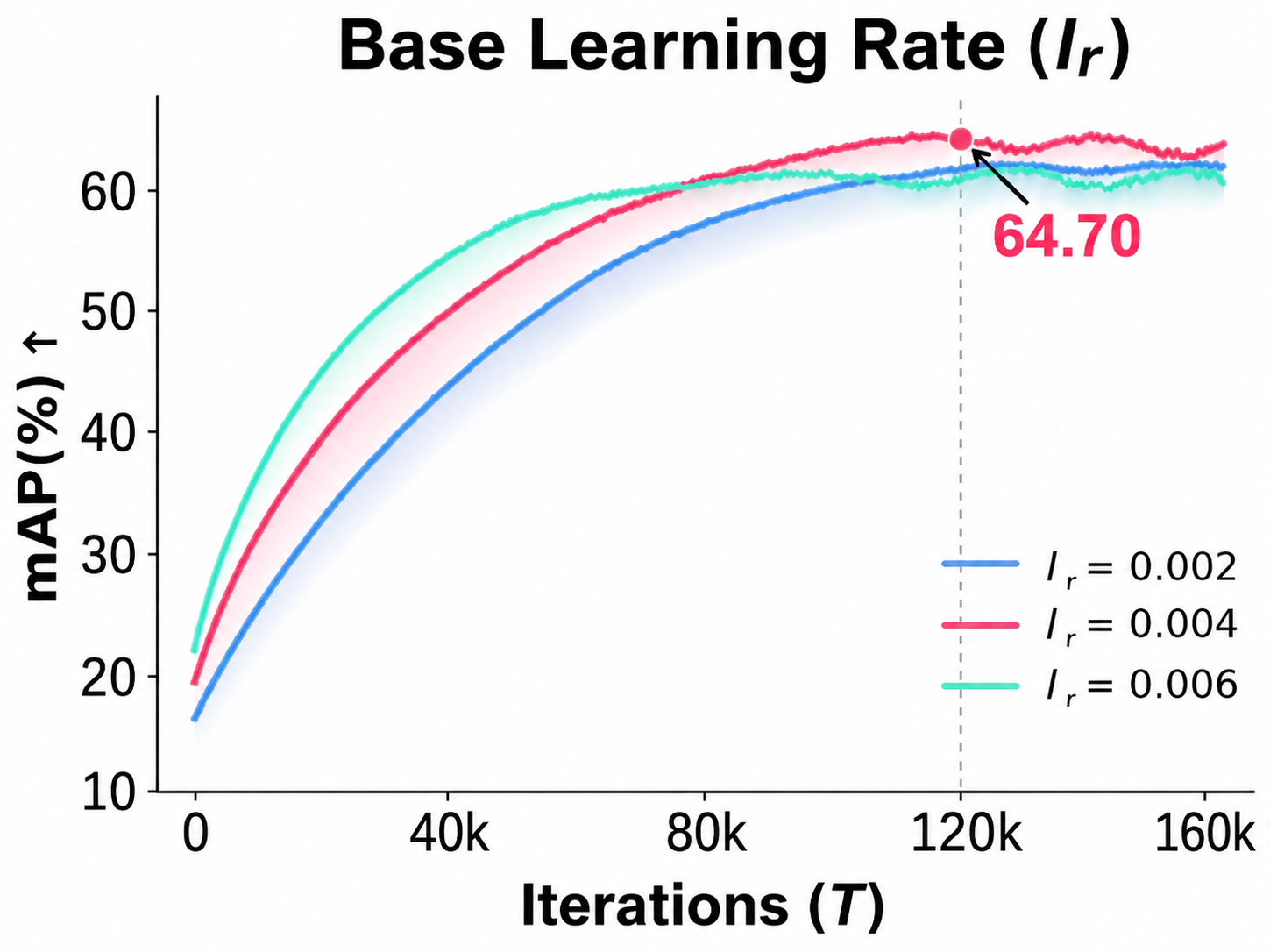}
        \captionsetup{skip=1pt}
        \captionof{figure}{Hyperparameter sensitivity analysis of the base learning rate ($l_r$).}
        \label{fig:lr_sensitivity}
    \end{minipage}

\end{figure}

% ================= Separate t-SNE Figure =================
\begin{figure}[t]
    \centering
    \includegraphics[width=0.88\linewidth]{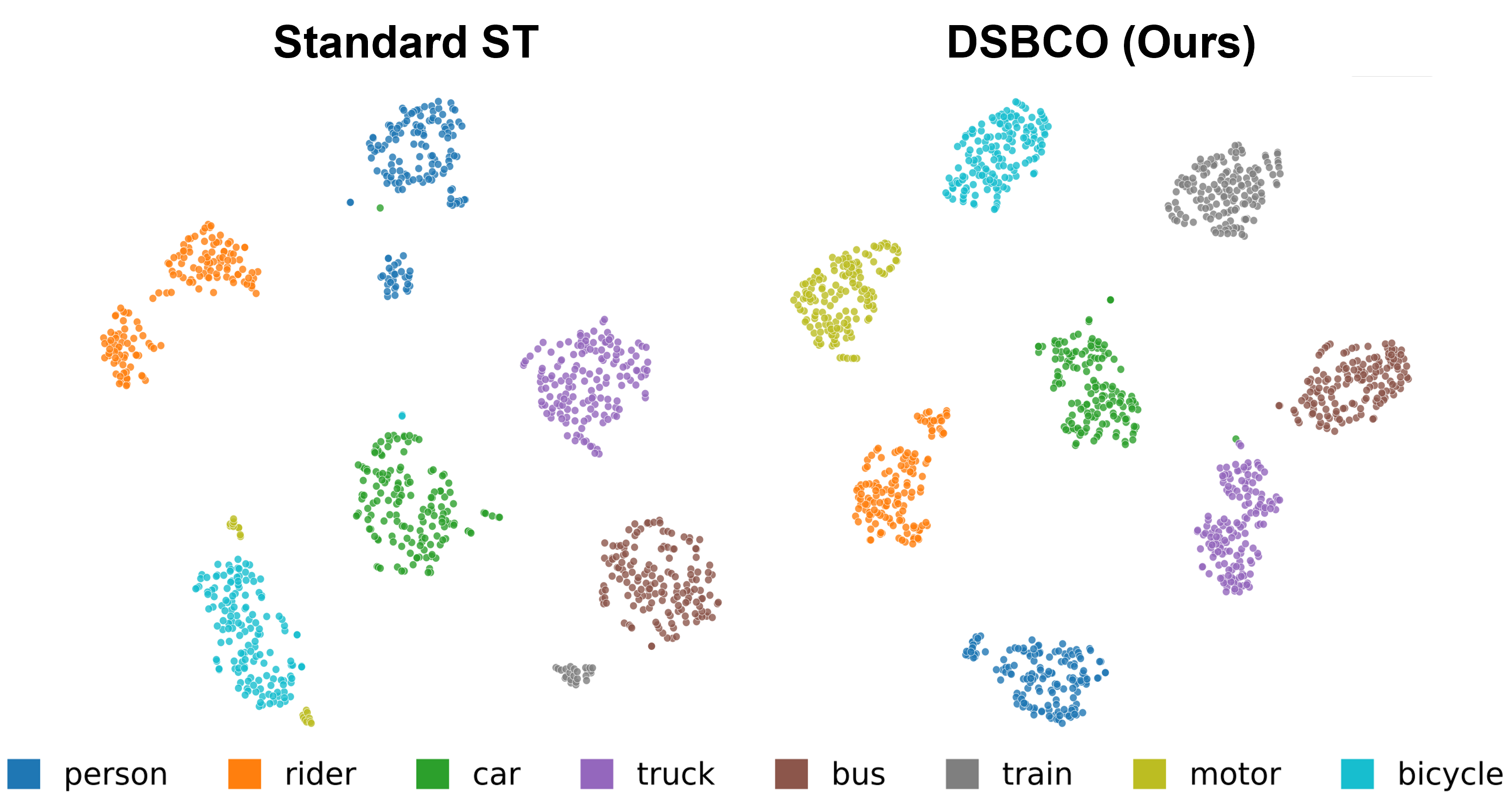}
    \caption{The t-SNE visualization of feature representations. Compared with Standard ST, DSBCO produces tighter clusters and clearer boundaries between different classes.}
    \label{fig:tsne}
\end{figure}

Quantitative evaluations across four cross domain scenarios demonstrate that DSBCO effectively transcends the generalization error bound. In Adverse Weather Adaptation (Table~\ref{tab:foggy}), DSBCO effectively achieves the best mAP of 64.7\%, outperforming the second-best overall method (MGCAMT) \cite{chen2025mgcamt} by 8.8\%, and exceeding the FCOS-based second-best (HT) \cite{deng2023harmonious} by 14.3\%. Significant gains in rigid structures across like Bus (72.3\%) confirm the regression cycle stream mitigates visibility-induced localization drift. In Diverse Context Adaptation (Table~\ref{tab:diverse_context}), our method establishes the best FCOS-based performance with 41.9\% mAP, improving approximately 2\% over HT under challenging scenarios. For Distinct Camera Adaptation (Table~\ref{tab:distinct_camera}), DSBCO attains a leading AP of 62.2\%, surpassing HT by approximately 2\% in challenging settings. Finally, in Synthetic Source Adaptation (Table~\ref{tab:synthetic_source}), DSBCO reaches the highest AP of 68.6\%, showing improvement over MGCAMT, while outperforming HT by 3.1\%. Overall, these results confirm that the bilevel optimization strategy successfully aligns distributions to prevent mode collapse results.

\subsection{Computational Efficiency}

Standard bilevel optimization is often too slow and unstable for object detection, as numerous repeatedly constructed large Hessian matrices from dense bounding-box samples can ultimately cause training collapse. Unlike the original CST approach \cite{liu21cycle}, DSBCO improves efficiency and scalability by using compact feature-based covariance matrix estimation instead of sample-based matrix inversion. This avoids substantial high memory costs and instability from large-matrix inversion. By using a linear solver and regression normalization, DSBCO skips heavy Hessian computations. As shown in Table~\ref{tab:time_cost}, DSBCO maintains training speeds similar to standard ST while ensuring robust convergence.

\subsection{Ablation Study}

Table~\ref{tab:ablation} evaluates DSBCO components on the Cityscapes $\to$ Foggy Cityscapes adaptation. The Standard ST baseline yields 41.2\% mAP. Adding Instance Consistency ($\mathcal{L}_{distill}$) provides an effective and stable improvement, boosting performance to 48.3\%. Building upon this, individually incorporating the Classification ($\mathcal{L}_{cstcls}$) or Regression ($\mathcal{L}_{cstreg}$) Cycle Stream elevates accuracy to 54.5\% and 50.9\%, respectively. Integrating all components jointly achieves the best 64.7\% mAP. As shown in Fig.~\ref{fig:hyperparam_sensitivity}, this optimal overall performance is reached using $\lambda_{cycle}=0.5$, $\lambda_{unsup}=0.01$, and $\eta=0.01$. Sensitivity results in Fig.~\ref{fig:lr_sensitivity} further show that DSBCO obtains the best performance at $l_r=0.004$. The complete hyperparameter analysis is provided in Supplementary \textbf{Section C}.

\subsection{Visualization Analysis}

Qualitative results are shown in Fig.~\ref{fig:vis_results}. DSBCO reduces false positives and improves detection across diverse scenes. Compared with the baseline, which often misses occluded objects or produces box misalignment, our framework yields more precise predictions with higher confidence. This indicates that DSBCO stabilizes training and enhances model robustness. Furthermore, Fig.~\ref{fig:tsne} visualizes feature embeddings via t-SNE~\cite{maaten08visualizing}. Compared with Standard ST, DSBCO yields compact clusters for Person and Rider, with clearer separations between Bicycle and Motor, and Train and Bus in embedding space.

\begin{figure}[t]
    \centering
    \includegraphics[width=0.88\linewidth]{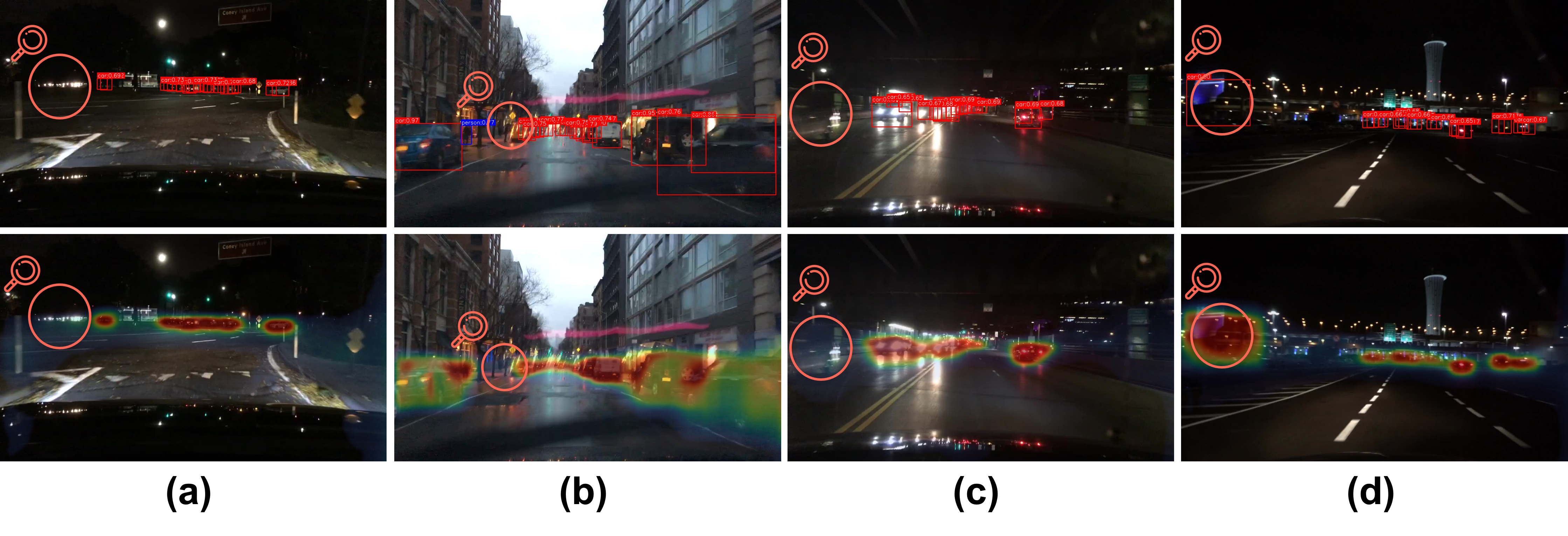}
    \caption{Case analysis of DSBCO under the Diverse Context Adaptation scenario.}
    \label{fig:failure_case}
\end{figure}

Finally, Fig.~\ref{fig:failure_case} presents failure cases and heatmaps under Diverse Context Adaptation. Case (a) shows false negatives for distant cars at night under low visibility, Case (b) shows missed nearby persons under daytime motion blur, Case (c) shows missed nearby cars during fast nighttime driving with severe blur, and Case (d) shows a night-time failure where Train is incorrectly classified as Car, indicating that these four failure cases require further investigation.

%% file: sec/5_conclusion.tex
\section{Conclusion and Limitations}
\label{sec:conclusion}
We propose DSBCO to overcome generalization barriers in domain adaptation 
by reformulating it as a bilevel optimization problem. DSBCO utilizes regres
sion normalization to effectively enforce cycle consistency and resolve regression instability. Empirical results confirm that DSBCO consistently outperforms competitive methods across diverse adaptation scenarios. However, several limitations still exist. First, extending this dual-stream optimization to modern architectures like Transformers requires further study. Second, the fixed thresholding mechanism may struggle against complex background noise in challenging scenarios, prompting us to explore more adaptive strategies in future work.